\documentclass[letterpaper, 10 pt, conference]{ieeeconf}  

\IEEEoverridecommandlockouts                              

\usepackage{graphics} 
\usepackage{epsfig} 
\usepackage{mathptmx} 
\usepackage{times} 
\usepackage{amsmath} 
\usepackage{amssymb}  

\usepackage{amsthm}
\usepackage{booktabs}
\usepackage{todonotes}
\usepackage{cleveref}
\usepackage{xcolor}
\usepackage[keeplastbox]{flushend}

\theoremstyle{definition}
\newtheorem{definition}{Definition}[section]

\title{\LARGE \bf
Exchangeable Input Representations for Reinforcement Learning
}

\author{John Mern, Dorsa Sadigh, and Mykel J. Kochenderfer
\thanks{The authors are with Stanford University, Stanford, California, 94305. Email: \{jmern91,dorsa,mykel\}@stanford.edu}%
}

\begin{document}

\maketitle
\thispagestyle{empty}
\pagestyle{empty}

\begin{abstract}


Poor sample efficiency is a major limitation of deep reinforcement learning in many domains.
This work  presents an attention-based method to project neural network inputs into an efficient representation space that is invariant under changes to input ordering. 
We show that our proposed representation results in an input space that is a factor of $m!$ smaller for inputs of $m$ objects. 
We also show that our method is able to represent inputs over variable numbers of objects. 
Our experiments demonstrate improvements in sample efficiency for policy gradient methods on a variety of tasks.
We show that our representation allows us to solve problems that are otherwise intractable when using na\"ive approaches.

\end{abstract}


\section{Introduction}

\begin{figure*}[!h]
\centering
\includegraphics[width=1.5\columnwidth]{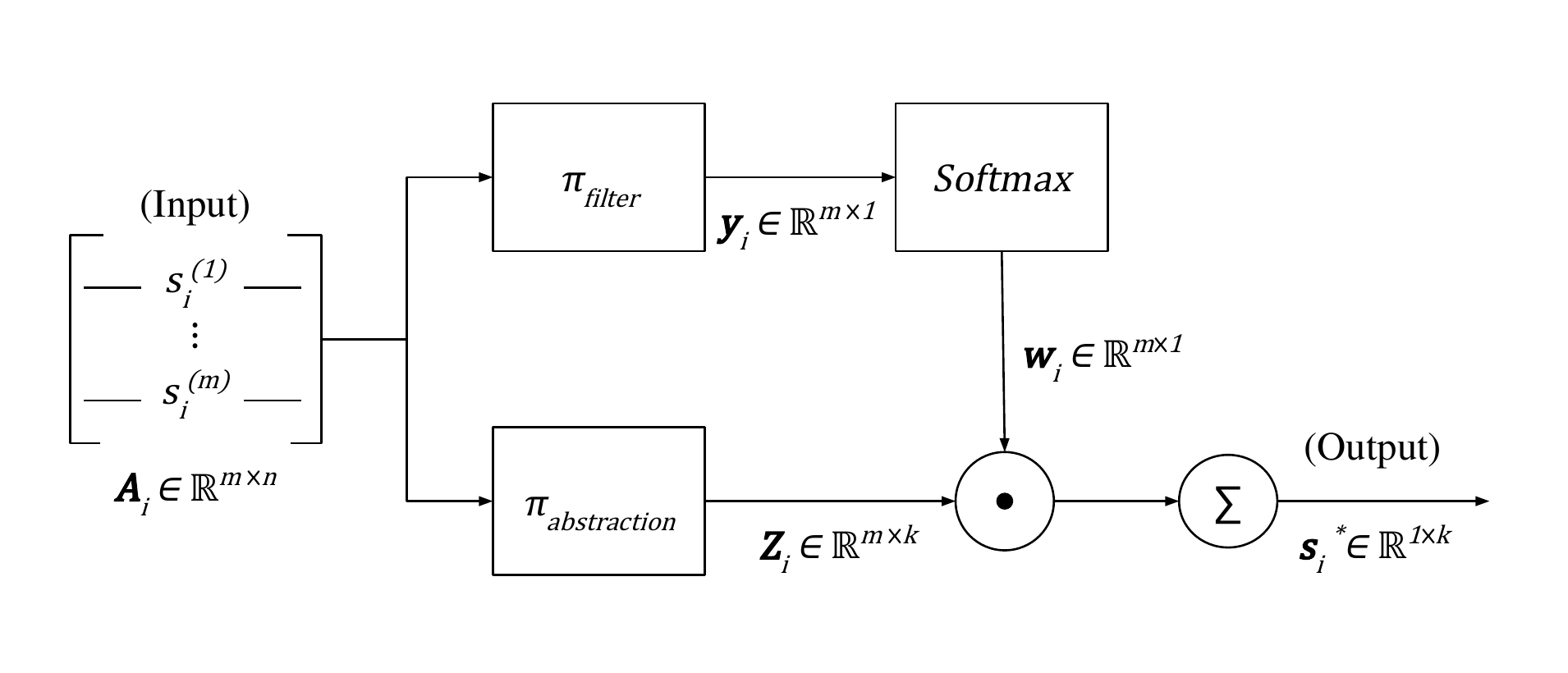}
\caption{Permutation invariant attention mechanism. Objects $\{s_i^{(1)}, \ldots, s_i^{(m)}\}$ of state $S_i$ are arrayed and passed into two neural networks $\pi_\text{filter}$ and $\pi_\text{abstraction}$. 
A softmax operation is performed on the $\pi_{filter}$ outputs and the resulting vector is element-wise multiplied with the outputs of $\pi_{abstraction}$.
This resulting array is summed along the $m$ dimension, resulting in the order-invariant output $s^*_i \in \mathbb{R}^k$.} 
\label{fig:attn}
\end{figure*}

Deep reinforcement learning (RL) achieves state-of-the-art performance across a variety of tasks~\cite{Silver2017}. 
However, successful deep RL training requires large amounts of sample data. 
Even relatively simple tasks can require millions to tens of millions of samples~\cite{Mnih2013}. 
While gathering large amounts of data in simulated domains may be achievable, it is often infeasible for planning and control of physical systems. 
Various learning methods have been proposed to improve sample efficiency. 
For example, model-based learning and incorporation of Bayesian priors use expert knowledge to reduce the data requirement~\cite{Gu2016, Spector2018}. 

The way that the input to an RL problem is represented can also impact the sample efficiency of a given learning approach. 
In multi-object environments, it is common to represent input states as concatenations of sub-state vectors of the objects within the environment. 
For example, a state in a robotic manipulation task may be represented as a set of the position and orientation vectors for all work pieces in the work space. 
In this case, we can refer to the sub-states of each piece as an \emph{object} in the factored state~\cite{mern2019}.

For many problems, an optimal policy should provide the same action for any permutation of the input set.
In these cases, the objects are \emph{exchangeable}. 
The key insight of this paper is that we can significantly improve efficiency by leveraging the exchangeable structure inherent in many reinforcement learning problems.

When inputs to neural networks are ordered sets, permutation invariance must be learned during training~\cite{Liu2019}. 
To avoid this additional learning requirement, methods have been proposed to represent inputs in an order-invariant form. 

The Object Oriented Markov Decision Process (OO-MDP) framework~\cite{Diuk2008} proposes such a method for exchangeable objects, however the presented methods are limited to discrete spaces with tabular representations. 
Approximately Optimal State Abstractions~\cite{Abel2016} proposes continuous approximations of the OO-MDP framework which extend to Q-learning problems with continuous input spaces.
Object-Focused Q-learning~\cite{Cobo2013} uses object classes to decompose the Q-function output space by interaction types, though it does not address exchangeability in the input space. 

This paper builds upon the insights presented in these works to propose a method to map any set of exchangeable objects to an order invariant representation. 
We show that applying such a mapping reduces the input space size by a factor of $m!$, where $m$ is the number of exchangeable objects.


Deep Sets~\cite{Zaheer2017} proposes a permutation invariant abstraction method similar to the one proposed in this paper. 
Additionally, they provide necessary and sufficient conditions for permutation invariant input mappings. 
Unlike our method, the method proposed produces a static mapping. 
That is, each input object is weighted equally in the invariant space regardless of value during the mapping. 

In contrast, our method proposes a permutation-invariant attention mechanism for the input mapping. 
Attention mechanisms are used in various deep learning tasks to dynamically filter the input to a down-stream neural network to emphasize the most important parts of the original input~\cite{xu2015,luong2015,jaderberg2015}. 
We adapt a dot-product neural network to efficiently apply dynamic attention~\cite{Vaswani2017}. 
We also propose a method to leverage the partial exchangeablity of environments with multiple object classes. 

An additional challenge facing deep RL comes from environments with varying numbers of objects. 
Most neural network architectures require a fixed input size. 
Tasks with varying numbers of objects are often solved with ad-hoc approaches such as input zero-padding. 
These methods can often lead to training inefficiencies~\cite{woof2018}. 
We show that the proposed attention mechanism can accept varying numbers of input objects without ad-hoc approximation. 

For review, the contributions presented in this paper are: 
\begin{itemize}
    \item An attention mechanism to map exchangeable object sets to permutation-invariant space.
    \item Demonstration that the attention mechanism is robust to varying numbers of input objects.
    \item A method to apply the attention mechanism to problems with multiple classes of exchangeable objects.
    \item Empirical study of the sample-efficiency gains of the abstraction method.
\end{itemize}

\section{Problem Statement}
Deep RL is a class of methods to solve sequential decision problems using deep neural networks.
To solve a sequential decision problem is to find a policy $\pi$ that maps state inputs to actions that maximize the expected discounted sum of rewards.  
Policies are closed-loop plans, in that they are reactive to the state of the environment.

It is common to represent the state of an RL problem as a set of objects. 
This approach leads the sample complexity of the problem to grow exponentially with the number of objects in the worst case~\cite{Robbel2016}.
This growth can be reduced by treating as \emph{exchangeable} the sub-state vectors corresponding to individual objects.
\begin{definition}
 We define a set of random variables $\mathcal{X}$ to be \emph{exchangeable} if and only if for any finite subset of $\mathcal{X}$, and any of the permutations of this subset, i.e., $\hat{\mathcal{X}}, \hat{\mathcal{X}}_\zeta \subset \mathcal{X}$: 
   \begin{displaymath}
    P(\hat{\mathcal{X}}) = P(\hat{\mathcal{X}}_\zeta),
  \end{displaymath}
  where $\hat{\mathcal{X}}$ is a finite set of random variables and $\hat{\mathcal{X}}_\zeta$ is an arbitrary permutation of $\hat{\mathcal{X}}$.
\end{definition}
Objects that may be treated as exchangeable are often referred to as being from the same \emph{class}. 
For example, in an autonomous vehicle control problem, we may consider automobiles as one class and pedestrians as another class.

Related to exchangeable variables are permutation invariant functions. 
Permutation invariant functions are functions that operate consistently over sets of exchangeable variables. 
In other words, the output of the function operating on a set of exchangeable variables is invariant under permutations of the set ordering. 
\begin{definition}
  If $S$ is a state defined by a set of objects and $\Pi$ is the set of all permutations on $S$, then a function $f$ is defined to be permutation invariant if and only if:
  \begin{displaymath}
    f(S) = f(\hat{S}) \ \forall \ \hat{S} \in \Pi 
  \end{displaymath}
\end{definition}

The problem this work seeks to address is to specify a permutation invariant function that can map ordered sets of exchangeable objects into an abstract state-space that 
\begin{enumerate}
    \item Retains all information necessary to solve the RL problem; 
    \item Can map sets of varying size; and
    \item Can be applied to problems with multiple object classes.
\end{enumerate}

\section{Proposed Approach}\label{sec:Method}
\begin{figure*}[t] 
\centering
\includegraphics[width=1.5\columnwidth]{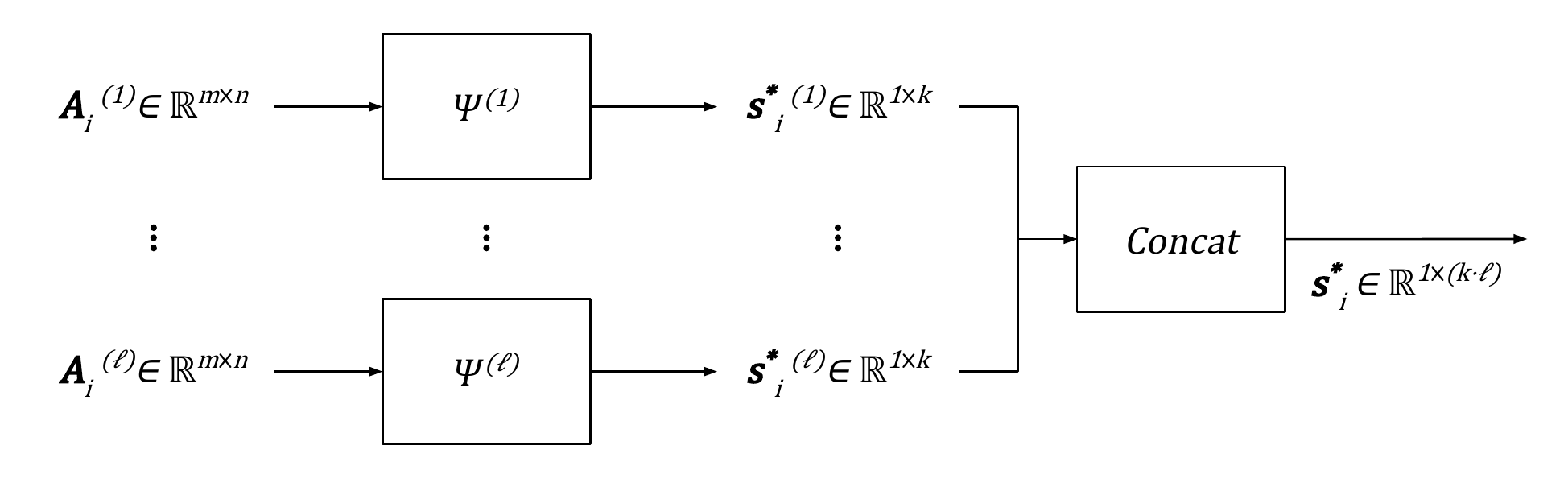}
\caption{Multi-class attention mechanism architecture. Object classes $S_i^{(j)}$ are arrayed to $A_i^{(j)}$ and passed into parallel attention mechanisms $\Psi^{(j)}$. The abstract output vectors $s_i^{(j)*}$ are concatenated into a single vector $s_i^* \in \mathbb{R}^{1\times(k*l)}$} 
\label{fig:multiclass}
\end{figure*}

Our objective is to reduce the sample complexity of deep reinforcement learning. 
To achieve this, we will leverage the exchangeablity of objects to reduce the size of the input space.
We will propose a method to map sets of objects $S_i$ to abstractions $S_i^*$ such that the mapping function is permutation invariant. 
Because abstracted space will be insensitive to order, the searchable input space of the reinforcement learning problem will be smaller, improving the sample efficiency of the learning process.


\noindent\textbf{Attention Mechanism.} 
We propose the attention network architecture shown in~\cref{fig:attn}, which is a permutation invariant implementation of dot-product attention. 
The mechanism is composed of two separate neural networks, abstraction network $\pi_{abstraction}$ and the filter network $\pi_{filter}$.
The abstraction network projects each object state vector into the permutation invariant space. 
The filter network generates an importance weight for each abstracted vector. 
Using the outputs of each, the abstracted state vectors are summed to produce the permutation invariant output $s^*_i$.

Assume the full state input $S_i$ is a set of $m$ object state vectors $\mathbf{s}_i^{(j)} \in \mathbb{R}^n$.
These state vectors are arranged in an array $\mathbf{A}_i \in \mathbb{R}^{m \times n}$.
The arrays are then passed through each fully-connected neural network.
The neural networks apply the transform $\sigma(\mathbf{X}_i^{(k)}W^{(k)} + \mathbf{b}^{(k)})$ at each layer, where $\sigma$ is a non-linear activation function, $\mathbf{X}_i^{(k)}$ is the input to the $k$ layer, and $W^{(k)}$ and $b^{(k)}$ are the weight and bias parameters respectively. 
Given this transform, each row of a layer's output is only dependent upon the corresponding row of the input. 

The $\pi_{filter}$ network outputs a weight for each input object as the vector $\mathbf{y}_i \in \mathbb{R}^{m \times 1}$.
The $\pi_{abstraction}$ network outputs an abstracted state for each object as the array $\mathbf{Z}_i \in \mathbb{R}^{m \times k}$.
The array and vector are multiplied element-wise, broadcasting along the $k$ dimension. 
The array is then summed along the $m$ dimension, resulting in the final abstracted state output $\mathbf{s}^*_{i} \in \mathbb{R}^k$.
It is the final summing operation that causes the mechanism to be permutation invariant.

The size of the final output $k$ is a design parameter, as are the particular designs of the neural networks. 
In this study, we found that setting $k \geq \hat{m} \times n$ provided good performance, where $\hat{m}$ is the average number of objects present. 
This allowed for full state information to be retained on average. 
The abstraction and filter neural networks used in this study were two-layer, fully connected networks with rectified linear units (ReLU) activations.

The parameters of the networks in the attention mechanism are not known a priori. 
They may be learned along with the parameters of the main graph. 
The sub-graph parameters may be updated using the gradient signal from the main loss term. 
No additional loss-term or gradient definition is required. 

\noindent\textbf{Sample Efficiency.} 
We can now define the search space reduction of an invariant mapping. 
Define a state space $\mathcal{S}$ such that $S = \{s_1, \ldots, s_m\}$ for $S \in \mathcal{S}$,
where $m$ is the number of objects. 
Let each object $s_i$ take on $n$ unique values. 
If we represent the states as sets of objects in the RL algorithm, then the state-space size $|\mathcal{S}|$ can be calculated from the expression for $m$ permutations of $n$ values. 

If all objects are exchangeable, there exists an abstraction that is permutation invariant. 
Since the order does not matter, the size of this abstract state $|\hat{\mathcal{S}}|$ can then be calculated from the expression for $m$ combinations of $n$ values:
\begin{equation}
|\mathcal{S}| = \frac{n!}{(n-m)!}, \ |\hat{\mathcal{S}}| = \frac{n!}{m!(n-m)!}
\end{equation}
Using a permutation invariant representation reduces the input space that the RL algorithm is required to search by a factor of $\frac{|S|}{|\hat{S}|} = \frac{1}{m!}$ compared to an ordered representation. 

\noindent\textbf{Permutation Invariance.} It can be shown that it is necessary and sufficient for a mapping $f$ to be invariant on all countable sets $\mathcal{X}$ if and only if it can be decomposed using transformations $\phi$ and $\rho$, where $\phi$ and $\rho$ are any vector valued functions, to the form~\cite{Zaheer2017}:
\begin{equation}\label{eq:invariant}
    f(X) = \rho\Big( \sum_{x\in\mathcal{X}} \phi(x)\Big)
\end{equation}
We will now demonstrate that the proposed mechanism can be factored into the permutation invariant form of~\cref{eq:invariant}. The variable names below correspond to the variable names in~\cref{fig:attn}:
\begin{align}
        s^*_i & = \sum_{j=1}^m z_i^{(j)}w_i^{(j)} \\
         & = \left(\sum_{j=1}^m e^{y_i^{(j)}}\right)^{-1}\sum_{j=1}^m z_i^{(j)}e^{y_i^{(j)}} \\ 
        \rho(X) & \leftarrow \left(\sum_{j=1}^m e^{y_i^{(j)}}\right)^{-1} \odot X\\ 
        x^*_i & = \rho\left(\sum_{j=1}^m \pi_{\text{inputs}}(s_i^{(j)}) e^{\pi_{\text{filter}}(s_i^{(j)})}\right)\\
        \phi(s_i^{(j)}) & \leftarrow \pi_{\text{inputs}}(s_i^{(j)}) e^{\pi_{\text{filter}}(s_i^{(j)})} \\
        s^*_i & = \rho\left(\sum_{j=1}^m\phi(s_i^{(j)})\right)
\end{align}
Equation $(4)$ follows from the definition of the Softmax operation.
The neural networks can be treated as functions operating on single objects because the parameters are shared for all objects.
The final line shows the mechanism in the form defined by equation $(2)$, completing the proof.

Additionally, it can be seen that the projection can handle variable numbers of input objects. 
As the number of individual state vectors $m$ changes, the dimension of the output vector $s_i^*$ remains constant as a result of the final summation operation. 
The dynamic weighting allows the network to focus on the retain more information about the important objects in the projection. 

\noindent\textbf{Multi-Class Attention.} 
In environments for which all objects are of the same class, the attention mechanism described can be applied directly. 
However, a slight extension is required for tasks with multiple object classes. 
Before defining that extension, we will provide a formal definition of class. 
\begin{definition}
  If $S$ is a state defined by a set of objects and $S^0$ and $S^1$ are disjoint subsets of $S$, $S^0$ and $S^1$ define object classes if all objects $s_i^0 \in S^0$ are exchangeable and $s_i^1 \in S^1$ are exchangeable.  
\end{definition}
In our robotic manipulator example, object classes could be defined by work piece type such as nuts and bolts.

In problems with multiple object classes, only objects within a given class are exchangeable. 
To address this, we can implement a separate attention mechanism for each object class.
The object vectors for each class $S_i^{(k)}$ are each arrayed and passed to a corresponding mechanism $\Psi^{(k)}$, each outputting a separate abstract state $s^*_{(k)}$, as shown in~\cref{fig:multiclass}. 

Each class-specific mechanism $\Psi^{(k)}$ has the same architecture as previously described and shown in~\cref{fig:attn}.
The outputs from each sub-graph are concatenated into a final abstracted input vector. 
Note that this input vector will be ordered; however, this is appropriate as the abstract class vectors are not exchangeable.

\section{Experiments}
We conducted a series of experiments to validate the effectiveness of our proposed abstraction. 
In the first two tasks a scavenger agent navigates a continuous two-dimensional world to find \emph{food} particles. 
The third task is a convoy protection task with variable numbers of objects. 

\begin{figure}
\centering
\includegraphics[width=0.9\columnwidth]{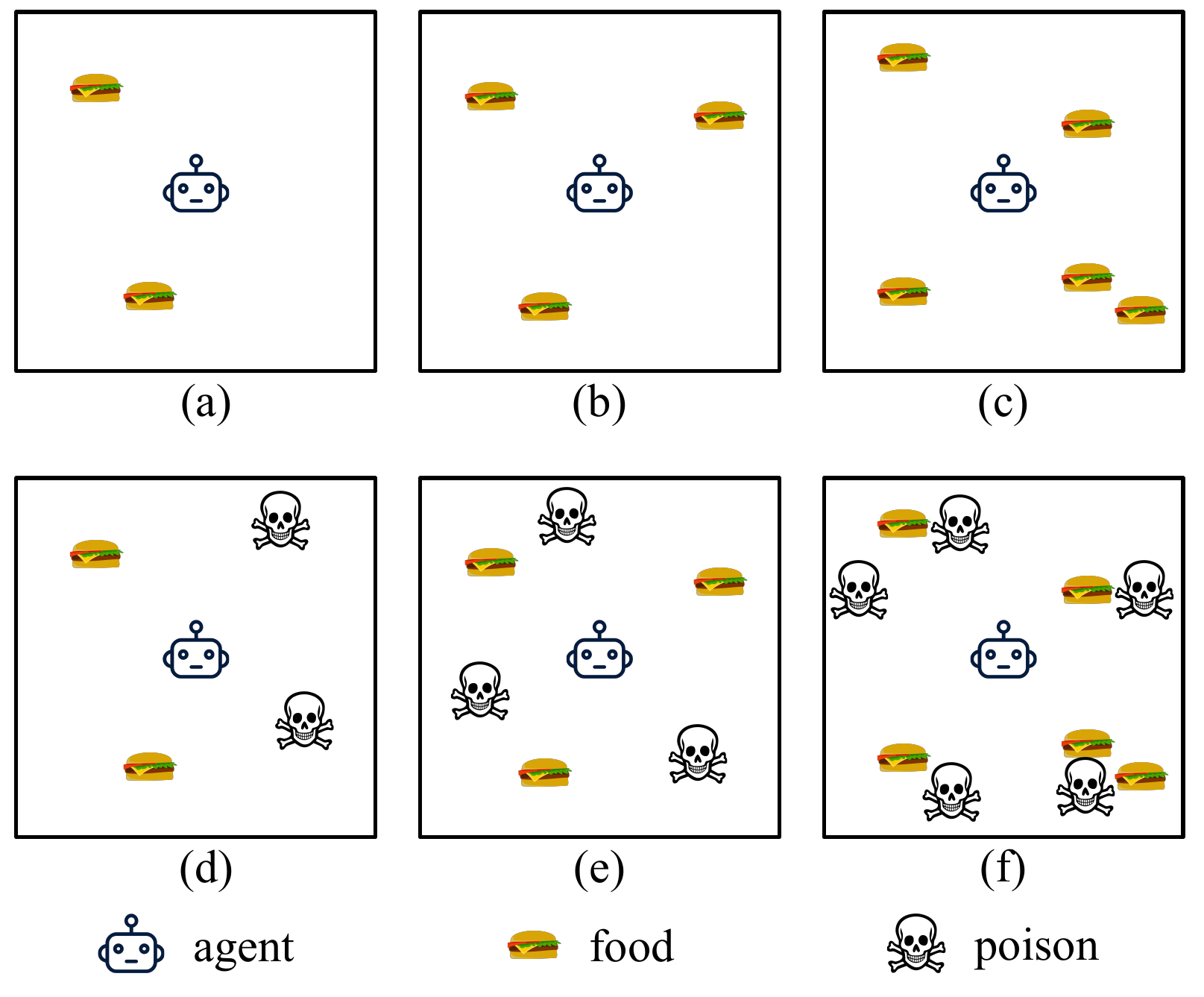}
\caption{Scavenger: (a) Task 1; 2 Objects (b) Task 1; 3 Objects (c) Task 1; 5 Objects (d) Task 2; 4 Objects (e) Task 2; 6 Objects (f) Task 2; 10 Objects} 
\label{fig:tasks}
\end{figure}

Scavenger Task 1 is designed to illustrate the effect of abstraction with a single class of objects.  
The goal of Scavenger Task 2 is to validate the effect of multi-class abstraction through the introduction of poison particles to the environment. 
Renderings of both of these environments are shown in~\cref{fig:tasks}.

\noindent\textbf{Task 1: Food Scavenger.}
The state space of Task 1 contains vectors $s \in \mathbb{R}^{2m+2}$, where $m$ is the number of target objects. 
The vector contains the relative position of each food particle as well as the ego position of the agent. 
The action space contains velocity vectors in two dimensions $a \in \mathbb{R}^2$ that are limited to a maximum velocity magnitude such that $\|a\| \leq a_\text{max}$. 
The agent receives a reward of $+1.0$ when reaching a food particle, and $-0.05$ for every time-step otherwise. 

The state transition model is deterministic with agent position updated at each time-step as shown below, where $\delta$ is the simulated time-step interval:
\begin{equation}
    s_{t+1} \leftarrow s_t + \delta a_t
\end{equation}
The agent is initialized at the center of the world at each episode and the food positions are sampled from a uniform distribution. 
The episode terminates upon reaching a food particle or when the number of time-steps exceeds a limit. 

We trained a stochastic policy to solve this task with and without the proposed attention mechanism. 
The baseline policy trained without our attention mechanism received a vector concatenation of the object state set, with each object's position in the vector remaining fixed through training. 
This is considered the standard RL approach. 
This same vector was used as an input to the attention mechanism, the output of which was used by the policy. 
All other training parameters were shared between the two approaches. 

The policies were feed-forward neural networks, with four hidden layers of 64 units each and Leaky ReLU activations. 
The network output parameters for a multivariate Gaussian distribution with diagonal covariance. 
The policy was trained using Proximal Policy Optimization (PPO)~\cite{Schulman2017}, with epoch batch sizes of 1,000 time steps and update batch-size of 256 steps. 
The policy ratio clipping parameter was set to 0.1 and no entropy bonus was provided. 
The reward signal used was advantage as calculated by the Generalized Advantage Estimation Lambda (GAE-$\lambda$)~\cite{Schulman2015}, with $\lambda=0.9$ and $\gamma=0.99$. 
Policies were trained for cases with varying numbers of objects, from two to five particles. 

\noindent \textbf{Task 2: Food Scavenger with Poisons.}
Scavenger Task 2 added one \emph{poison} particle for each food particle in the environment. 
If an agent reaches a poison particle, a reward of $-1.0$ is given and the episode terminates. 
As with the food particles, the initial positions of the poison particles are sampled from a uniform distribution. 
The remainder of the task is identical to Task 1.
As before, we train our policy with a baseline and abstracted representation. 

This extended task was developed to test the effect of abstraction over multiple classes. 
In our framework, poison objects are in a separate class from food objects. 

\noindent\textbf{Task 3: Convoy Protection.}
A final experiment was conducted on a more difficult task in which the number of objects varies across episodes. 
The task requires a defender agent to protect a convoy that follows a predetermined path through a 2D environment. 
Attackers are spawned at the periphery of the environment during the episode, and the defender must block them while they attempt to approach the convoy. 
The environment was simulated in Anvel, a high-fidelity ground-vehicle simulation engine. 
\begin{figure*}[!h]
\centering
\includegraphics[width=1.9\columnwidth]{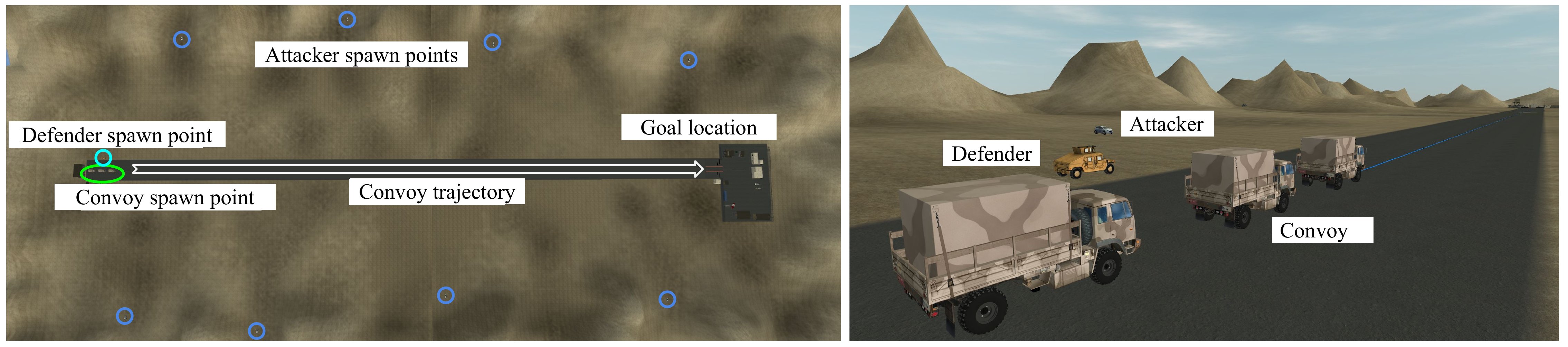}
\caption{Robotic Convoy Task Environment. Defender agent must protect the convoy as it travels across the environment by blocking the attackers from approaching.} 
\label{fig:convoy}
\end{figure*}
We decomposed the solving of this problem using a hierarchical learning approach in which a primitive policy was trained a priori and used with fixed parameters over the training of a high-level policy. 
The primitive policy was trained to map the desired vehicle location change to vehicle commands (wheel orientation and throttle). 
Each time-step for the primitive policy was set to 0.1 second of simulation world time.  

Because the training of the high-level policy is of interest to this work, we will define the task in terms of its inputs and outputs.
The state space is the space of vectors representing the state of each non-ego vehicle in the environment $(x, y, status)$, where the $status$ is a binary flag of whether or not the object is currently active. 
The full state also contains the state of the ego-vehicle $(x, y, \theta)$, where $\theta$ is the z-axis orientation of the vehicle. 
The action space space contains vectors of the reachable changes in position. 
The three vehicle convoy was generated at a fixed point at the left side of the environment for each episode and traveled at a constant rate toward the right. 
The attackers were spawned at random times from one of eight spawn points at the periphery of the environment. 
The attackers approached the closest convoy member and with maximum speed equal to twice the convoy speed. 

The episode terminates when all convoy members either reach the goal position or are reached by an attacker. 
The agent receives a reward of $-1.0$ for each convoy member that is attacked and a reward of $+0.1$ for each attacker that is successfully blocked. 
As with previous experiments, we trained this policy with a baseline representation in which all object states were concatenated in fixed-order vectors and with a representation generated with our proposed methods. 

\section{Results}
\begin{figure}[!t]
    \centering
    \includegraphics[width=0.9 \columnwidth]{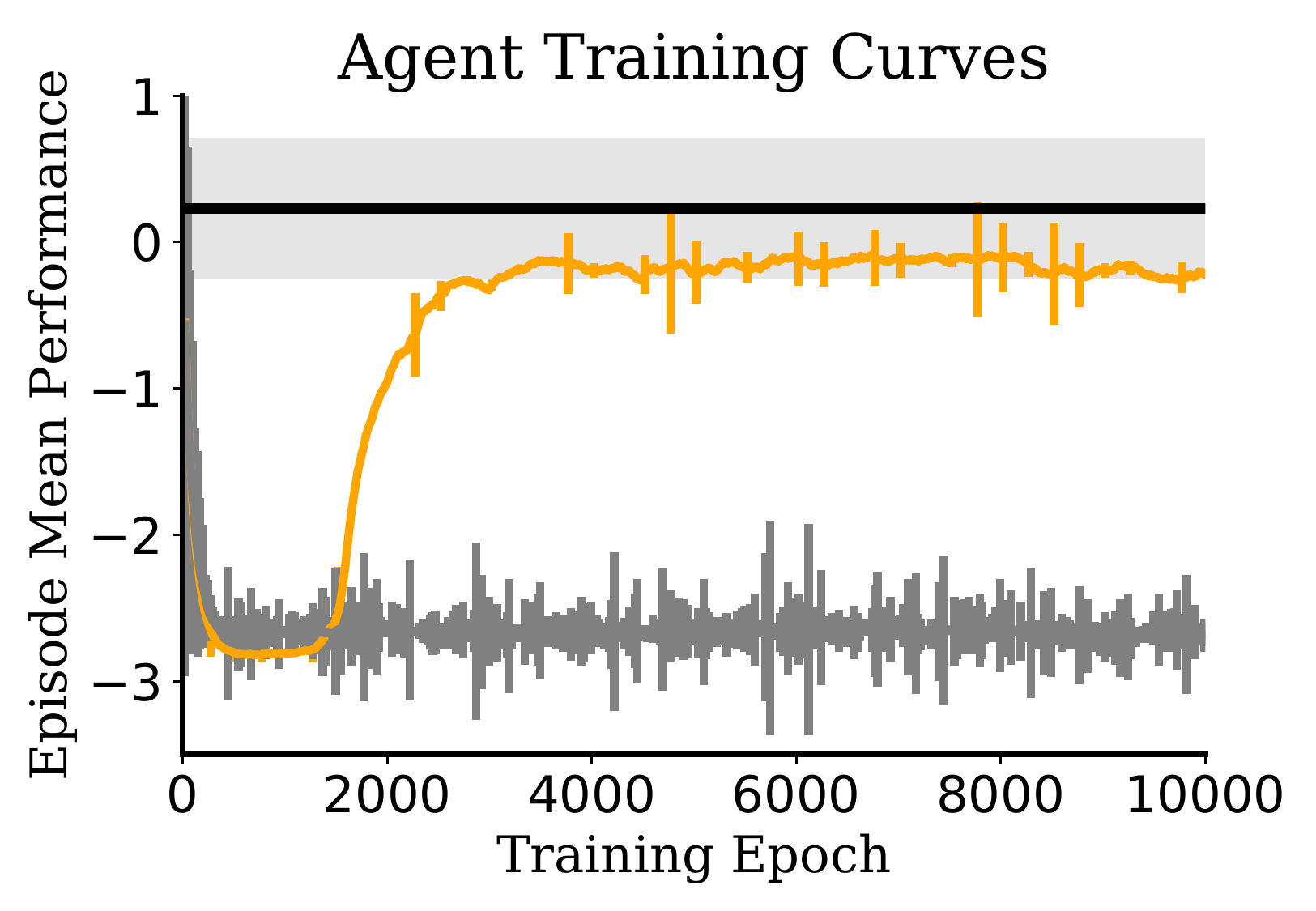}
    \includegraphics[width=0.9 \columnwidth]{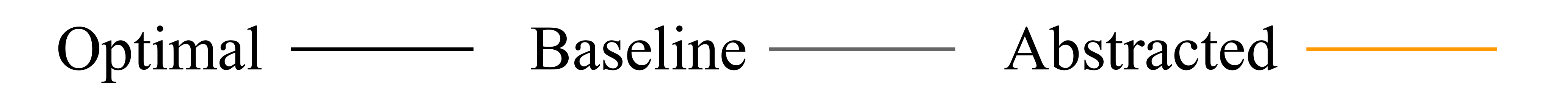}
    \caption{Convoy Task Training Curve. ``Abstracted Score" shows performance of agent with attention mechanism. ``Baseline Score" shows performance of the agent without attention mechanism.}
    \label{fig:ConvoyTraining}
\end{figure}

\begin{figure*}[!h]
    \centering
    \includegraphics[width=0.5\columnwidth]{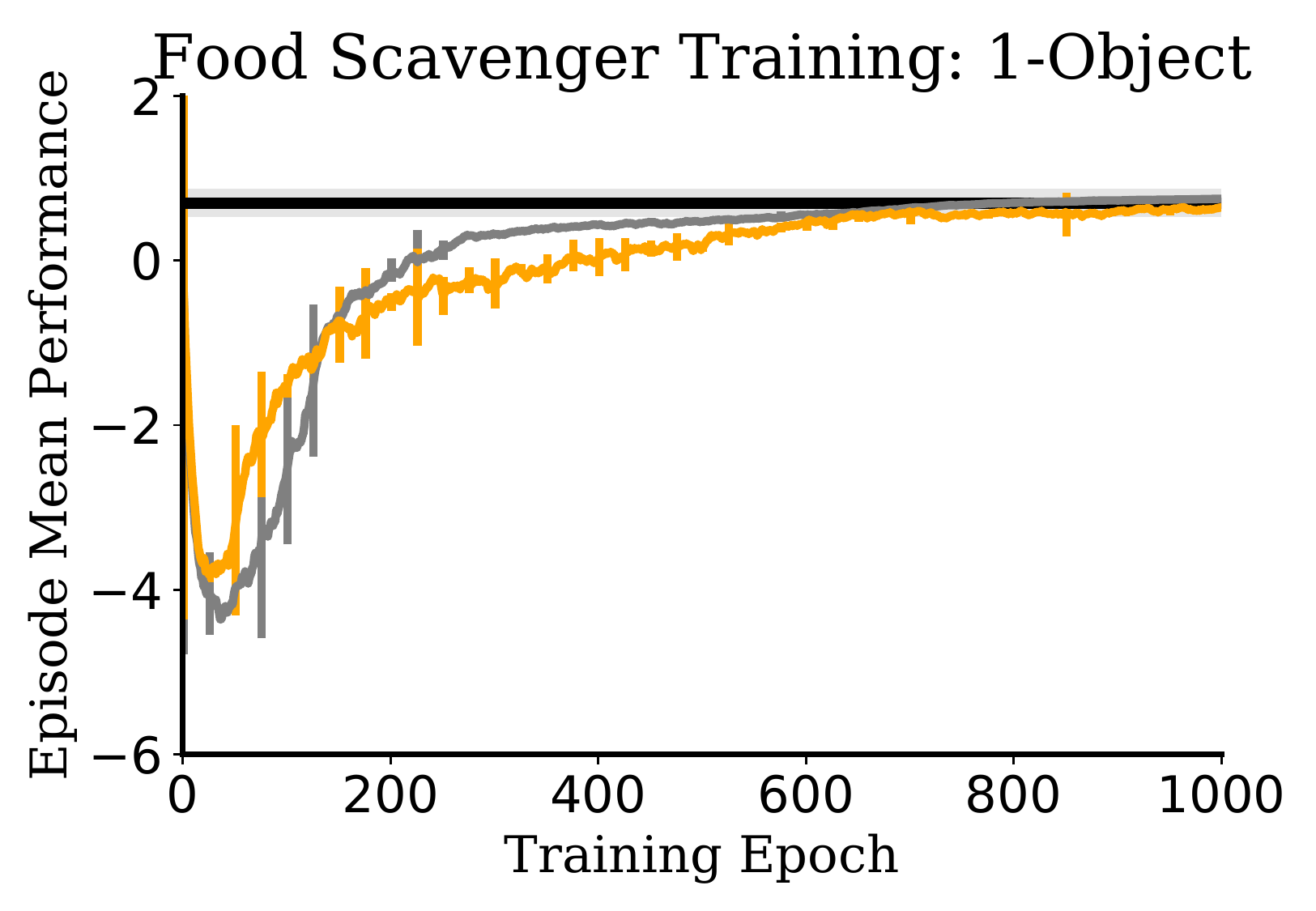}
    \includegraphics[width=0.5\columnwidth]{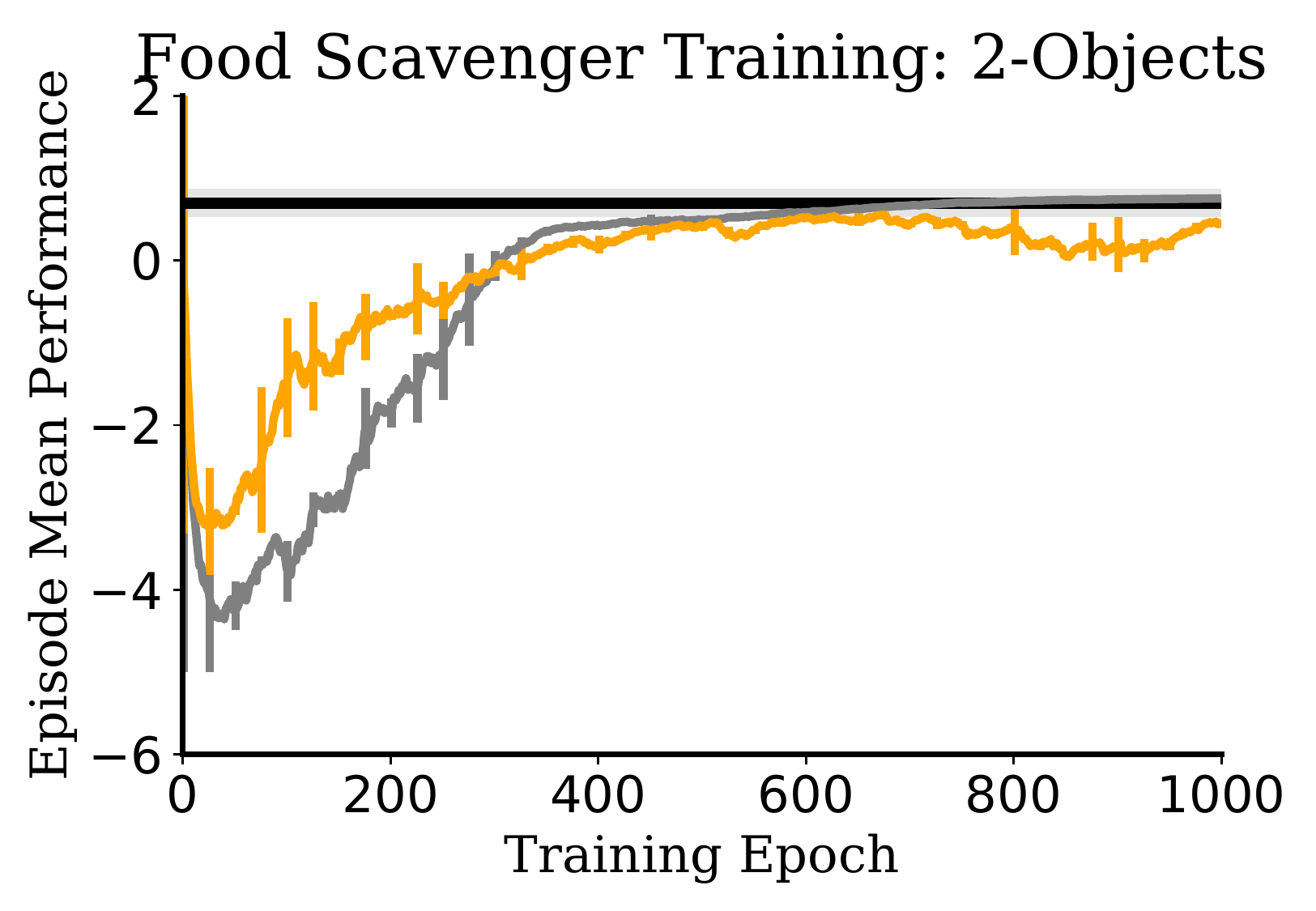}
    \includegraphics[width=0.5\columnwidth]{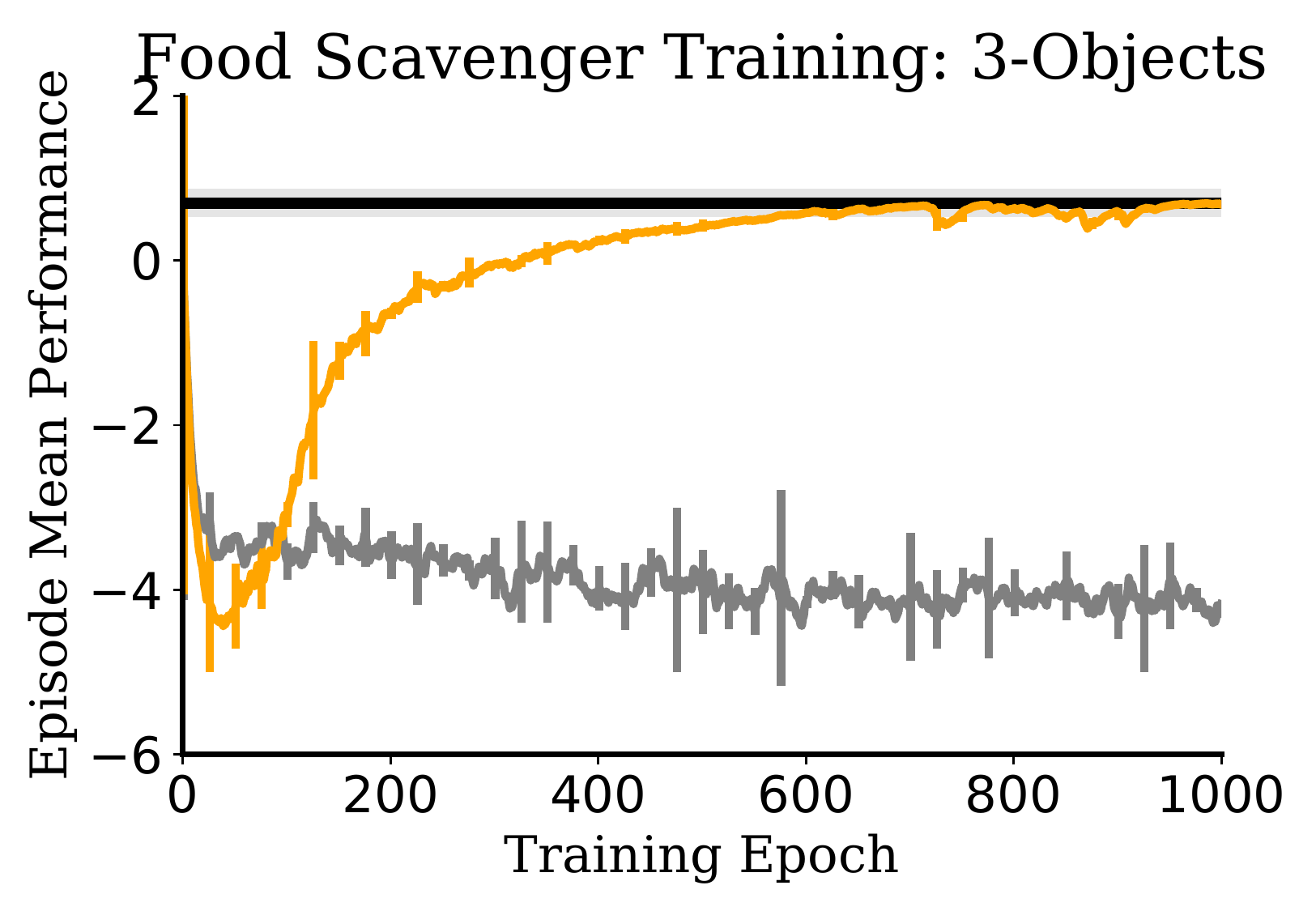}
    \includegraphics[width=0.5\columnwidth]{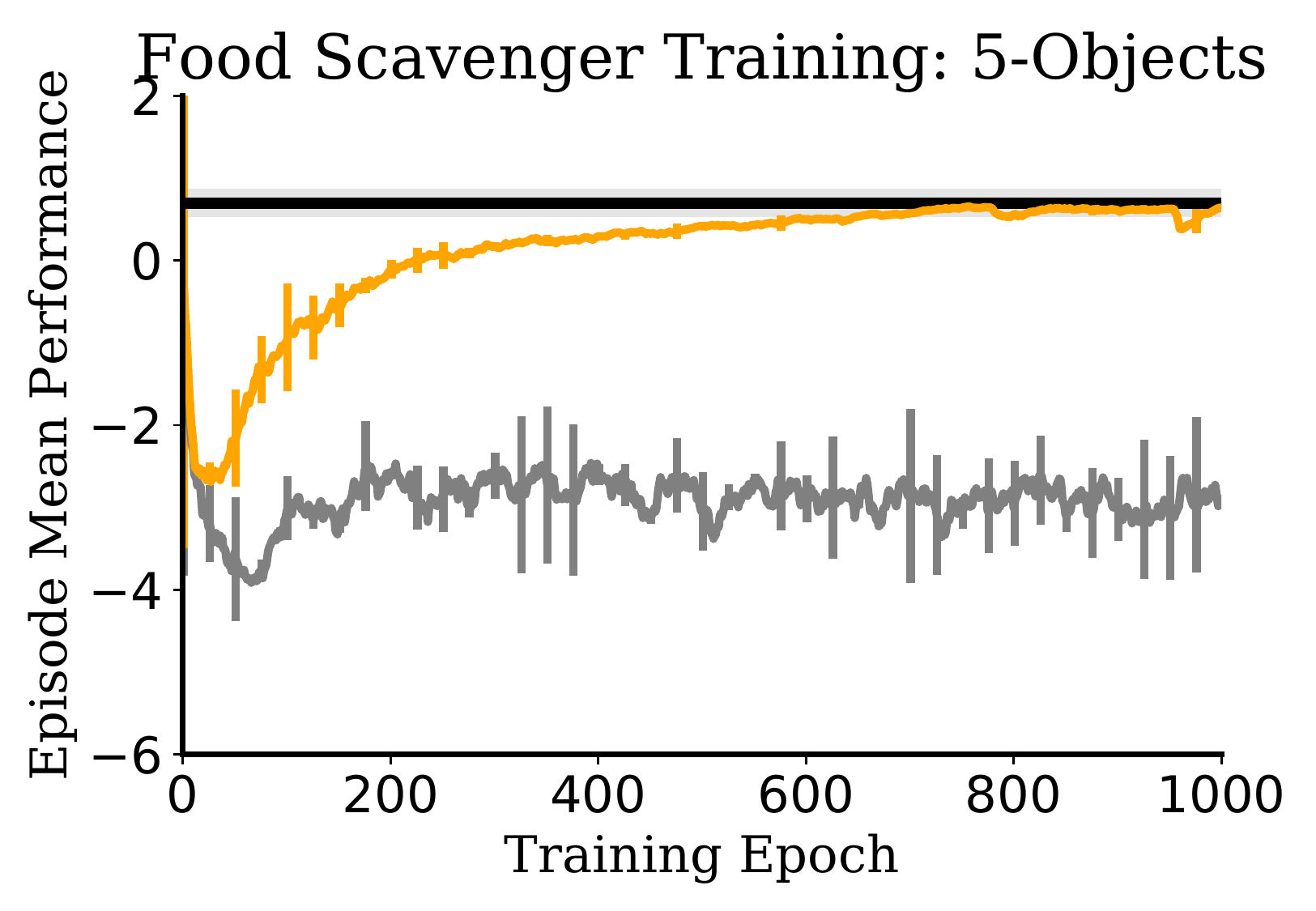}
    \includegraphics[width=0.9\columnwidth]{figs/ConvoyLegend.pdf}
    \caption{Scavenger Task 1 Training Curves. Each graph shows learning on task with given number of target objects.}
    \label{fig:SATraining}
\end{figure*}

\begin{figure*}[!h]
    \centering
    \includegraphics[width=0.5\columnwidth]{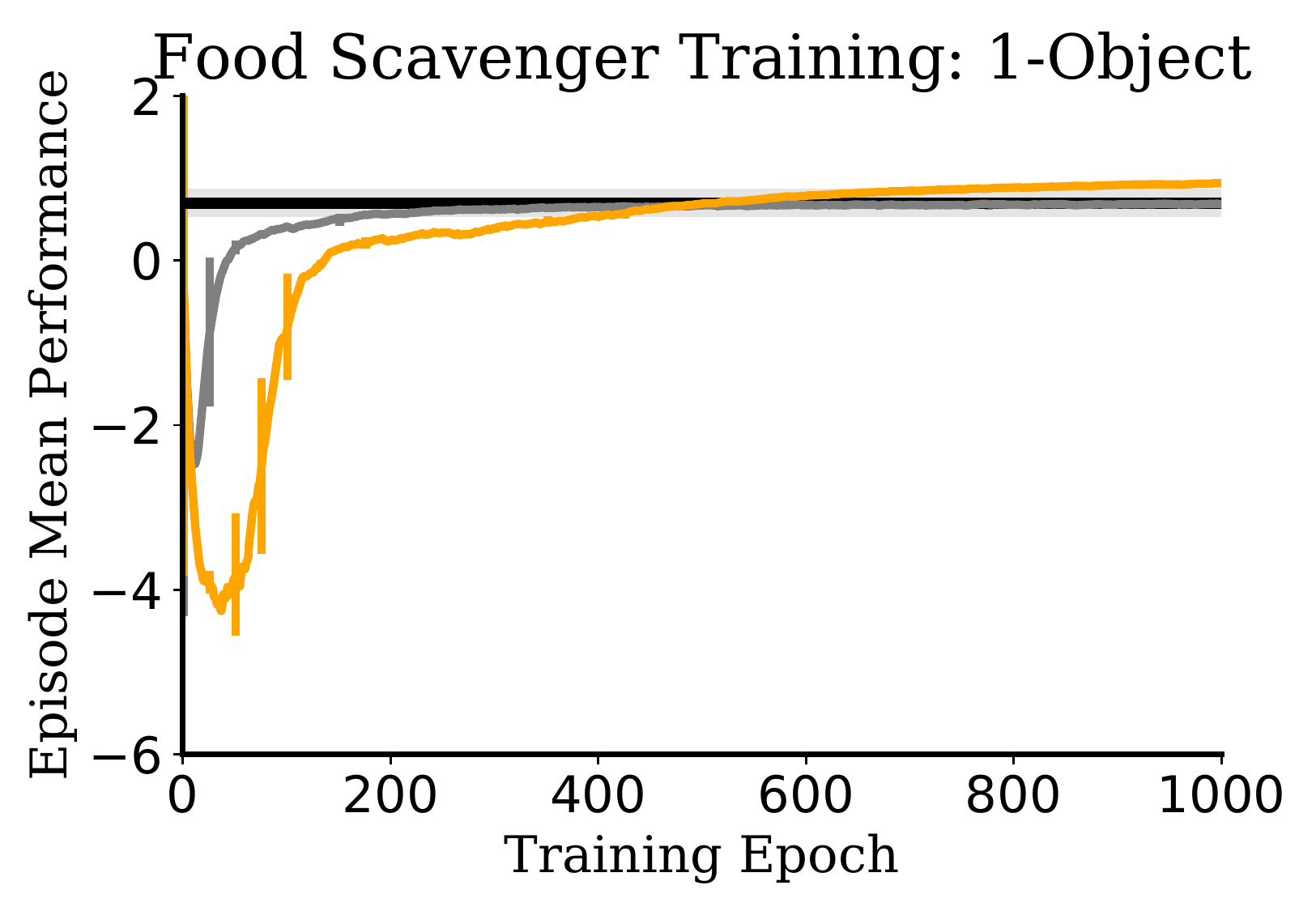}
    \includegraphics[width=0.5\columnwidth]{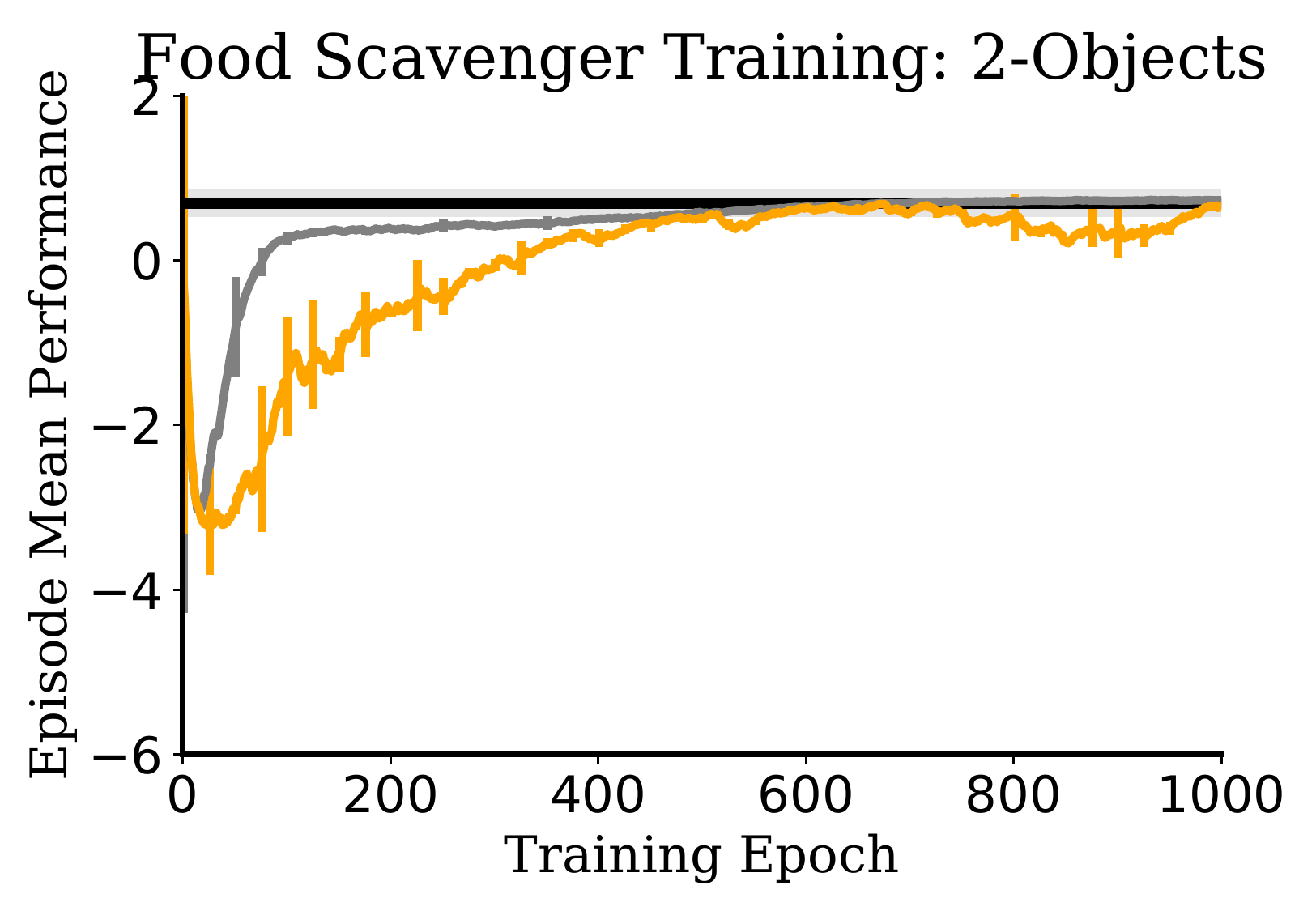}
    \includegraphics[width=0.5\columnwidth]{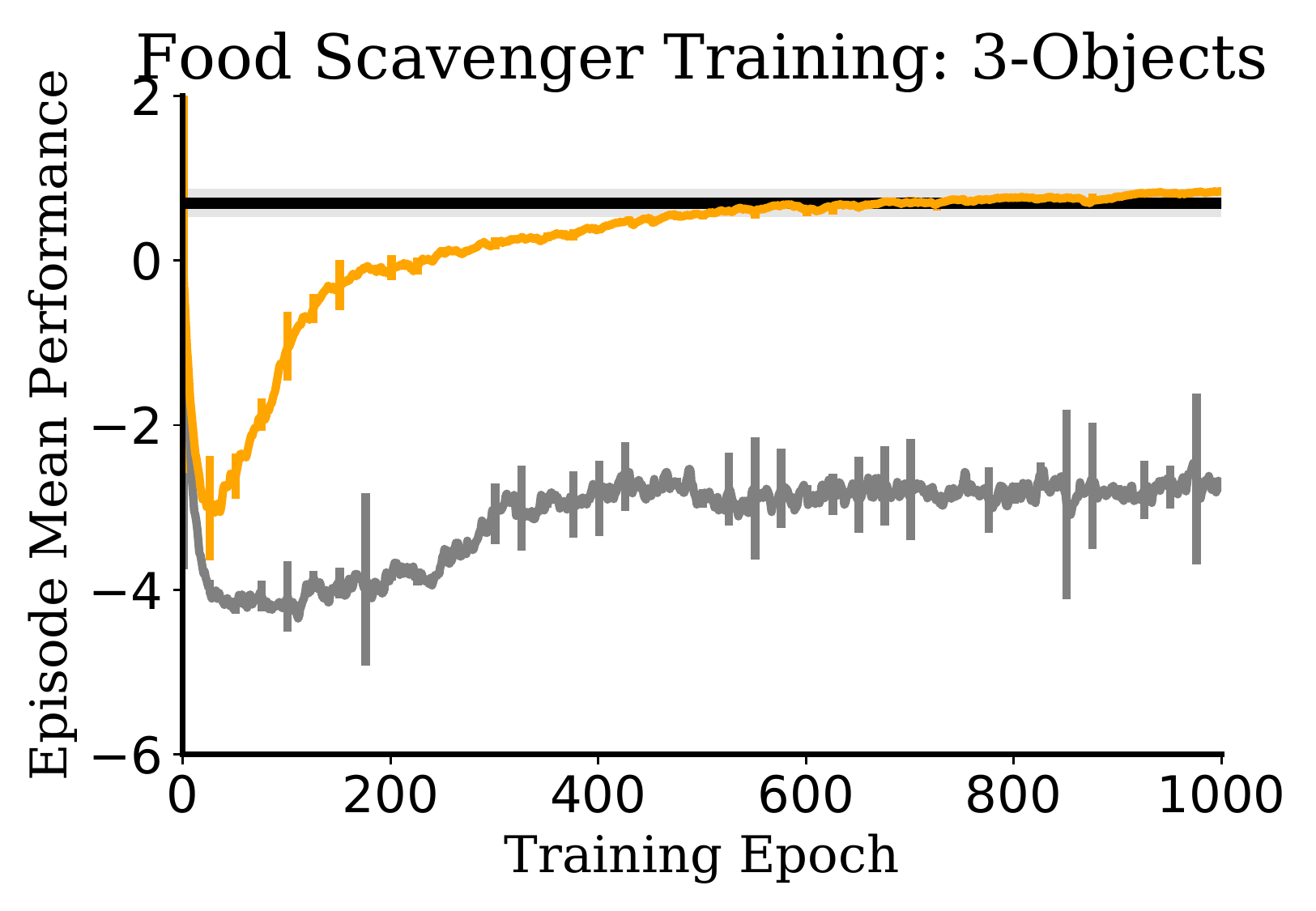}
    \includegraphics[width=0.5\columnwidth]{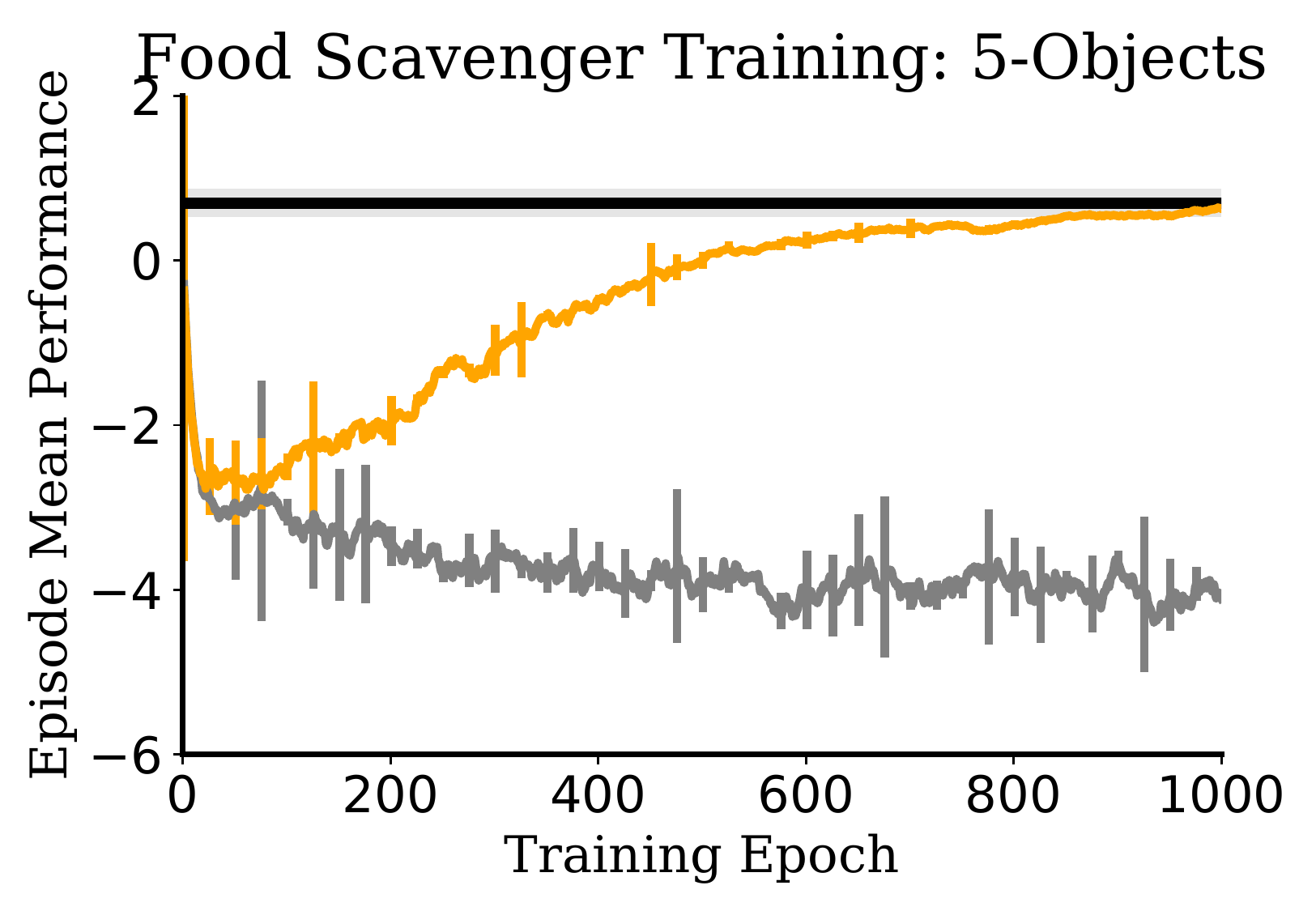}
    \includegraphics[width=0.9\columnwidth]{figs/ConvoyLegend.pdf}
    \caption{Scavenger Task 2 Training Curves --- Each graph shows learning on task with given number of target objects.}
    \label{fig:OATraining}
\end{figure*}

The Scavenger 1 training curves for the baseline and abstracted policies are shown in~\cref{fig:SATraining}. 
A simple optimal policy was defined for the task (travel toward to closest particle) and the performance of this policy is also shown on the graphs.

The introduction of the permutation invariant representation allowed the RL algorithm to efficiently scale to tasks with more objects. 
PPO failed to solve the problem in 1,000 training epochs using the naive representation for more than two targets. 
Using the abstraction, PPO solved the problem for tasks up to five targets. 
A slight slowdown in early learning can be seen in the abstracted cases, likely due to the need to learn the parameters of the abstraction sub-graph. 

In Scavenger Task 2, PPO fails to scale beyond two targets (four total objects) while learning on the naive representation. While using our proposed permutation invariant representation, the algorithm effectively scales to cases of up to 10 total objects in only 1,000 training epochs.  

The abstraction was also tested for the more difficult convoy protection problem. 
This problem presented the additional challenge of accommodating an input space with a variable number of objects and a long time horizon.
In this case, PPO was completely unable to learn using the naive representation over 10,000 training epochs containing 20M sample time steps. 
With the invariant abstraction, the algorithm was able to successfully learn a policy that achieved average performance within the one-$\sigma$ bound of the optimal policy in only 3,000 epochs (6M time steps). 
This demonstrates that even in complex tasks, a significant improvement in sample efficiency is gained with the abstraction. 

\section{Conclusion}
\noindent\textbf{Summary.}
We presented an attention-based method to project sets of object state vectors into representations that leverage the exchangeability of objects. 
We showed that this attention mechanism is permutation invariant. 
In order to apply this mechanism across multiple object classes, we presented a simple extension, allowing the leveraging of class-dependent object exchangeability. 
The proposed mechanism was also shown to accommodate varying numbers of objects. 
We demonstrated the effectiveness of the approach to enhance the scalability of the PPO policy gradient learning algorithm on a set of demonstration problems. 
Our simple scavenger tasks highlighted the effect of ignoring exchangeability, with PPO unable to scale with the number of objects. 
In addition, we demonstrated the effect of using the method in a more difficult hierarchical learning problem.


\noindent \textbf{Limitations and Future Work.} 
Though the abstraction improved the ability to learn with more objects, it was observed to slow down learning on tasks with fewer objects. 
This is likely due to the need to learn the additional parameters of the attention network. 
For tasks with fewer objects, a static mapping such as proposed in Deep Sets may be appropriate. 

Another limitation of our work is the reliance on dot-product attention. 
While this method does enable dynamic weighting of objects during mapping, the importance is determined for each object without considering the other objects present. 
An attention mechanism that also considers interactions between objects in weighting could provide better performance. 

Our work only addressed the effects of exchangeability in state representation, though similar investigation should be made into other parts of the problem. 
In particular, extensions of the concept of interaction classes should be developed to leverage exchangeability in the action space and transition function.
In our approach, we fixed several elements of the attention graph, such as the abstracted state dimension, as hyperparameters. 
The effect of these on learning rate and converged policy performance should be further investigated.



\bibliographystyle{IEEEtran} 
\bibliography{bibliography}






\end{document}